\documentclass[graybox]{svmult}

\usepackage{type1cm}        
\usepackage{makeidx}         
\usepackage{graphicx}        
\usepackage{multicol}        
\usepackage[bottom]{footmisc}

\usepackage[linesnumbered,ruled,vlined]{algorithm2e}
\usepackage{algorithmic}

\usepackage{amssymb}
\usepackage{amsmath}
\usepackage{amsfonts}
\usepackage{enumerate}

\usepackage{newtxtext}       %
\usepackage{newtxmath}       

\usepackage{comment}

\newcommand{\bx}{\mathbf{x}} 
\newcommand{\bob}{\mathbf{f}} \newcommand{\ks}{\mathit{k_{\text{slow}}}}

\makeindex             

\begin{document}

\title*{Heterogeneous Objectives: State-of-the-Art and Future Research }
\author{Richard Allmendinger and Joshua Knowles}
\institute{Richard Allmendinger \at Alliance Manchester Business School, The University of Manchester, Manchester, UK \email{richard.allmendinger@manchester.ac.uk}
\and Joshua Knowles \at Invenia Labs, Cambridge, UK\\Hon. Professor, Alliance Manchester Business School, The University of Manchester, UK\\ \email{joshua.knowles@invenialabs.co.uk}}
%
%
\maketitle

\abstract{Multiobjective optimization problems with \emph{heterogeneous objectives}\/ are defined as those that possess significantly different types of objective function components (not just incommensurable in units or scale). For example, in a heterogeneous problem the objective function components may differ in formal computational complexity, practical evaluation effort (time, costs, or resources), determinism (stochastic vs deterministic), or some combination of all three.  A particularly challenging variety of heterogeneity may occur by the combination of a time-consuming laboratory-based objective with other objectives that are evaluated using faster computer-based calculations. Perhaps more commonly, all objectives may be evaluated computationally, but some may require a lengthy simulation process while others are computed from a relatively simple closed-form calculation. In this chapter, we motivate the need for more work on the topic of heterogeneous objectives (with reference to real-world examples), expand on a basic taxonomy of heterogeneity types, and review the state of the art in tackling these problems. We give special attention to heterogeneity in evaluation time (latency) as this requires sophisticated approaches. We also present original experimental work on estimating the amount of heterogeneity in evaluation time expected in \emph{many}-objective problems, given reasonable assumptions, and survey related research threads that could contribute to this area in future.}

\section{Motivation and Overview}\label{sec:Motivation}
It is a familiar thing to anyone who works in optimization that the objective functions in different optimization problems vary along important `dimensions', giving rise to different large classes of problems. For example, we are aware that objectives may take discrete, continuous or mixed input variables; they may give bounded or unbounded outputs which may be discrete or continuous; they may be closed-form expressions or require more-or-less complicated simulations; if they are closed-form they may be linear, nonlinear but convex, or nonconvex; they may be non-computable in practice, instead requiring experimental processes to be evaluated; and, they may be precise, certain and repeatable, or they may be uncertain objective functions, subject to parametric or output noise. 

Given this natural variety in the objective functions seen in different (single-objective) problems, it should not be surprising that when considering multiobjective optimization problems, the different objective function components forming the overall function may also be quite different from one another. Indeed, one aspect of the difference is very well known and well accounted for historically: the function components may give outputs in different units, and those units may be incommensurable; this incommensurability of objective function values is a key motivation of ``true'' multiobjective methods --- those that do not form combinations of objective values, except with explicit reference to a proper preference model or preference elicitation process.

Although it is not surprising that different objective functions should be not just different in the scale, range or units of their outputs (incommensurable), but different in the more fundamental ways listed above, there has been very little stated explicitly about this heterogeneity in the literature, and almost no work that seeks to offer ways to handle it inside multiobjective algorithms. The first work the authors are aware of was by us~\cite{allmendinger2013hang}.

We were motivated in that work to consider a particular type of heterogeneity that could potentially cause a lot of inefficiency in a standard evolutionary multi-objective optimization (EMO) approach, namely that the objective function components were of different {\em latency}\/ i.e., evaluation time. Our motivation came from a number of real problems we had been looking at under the banner of ``closed-loop optimization'', which is to say at least one of the objective function components was dependent on a non-computational experiment for evaluation, such as a physical, chemical or biological experiment. Allmendinger's PhD thesis~\cite{Allmendinger2012phd} studied many different consequences apparent in these closed-loop problems (including dynamic constraints and interruptions), and this followed much real experimental work in this vein by Knowles and co-authors~\cite{Kno2009closed,ohagan2005closed,ohagan2007closed,platt2009aptamer}. Varying per-objective latencies can, of course, arise also in scenarios requiring intense computational experiments, such as CFD and other complex simulations; for example, obtaining objective function values may require the execution of multiple time-consuming simulations, which may vary in running time.

Following our publication of the above-mentioned paper~\cite{allmendinger2013hang}, which dealt with heterogeneity in latency between the different objectives of a problem, Knowles proposed a broader discussion topic on Heterogeneous Functions~\cite{eichfelder4} at a Dagstuhl Seminar~\cite{Dagstuhl15031}, in which the topic was fleshed out by the participants of the seminar over the course of three or four days. Since then, there have been two significant strands of work. First, we (the authors of this chapter) have presented a further study on heterogeneity~\cite{ejor:Latencies}, in which we extended our work on handling differing latency within a pair of objectives in a bi-objective problem. This was also further extended in~\cite{chugh2018surrogate}, and then by Jin and co-authors in~\cite{10.1145/3377930.3390147}, with both of these studies adapting surrogate-assisted evolutionary algorithms to cope with latencies between objectives. Secondly, Thomann and co-authors have published a pair of papers, and PhD thesis, on heterogeneity~\cite{thomann2019representation,thomann2019trust,thomann2019trust_PhD} particularly concerning non-evolutionary methods.

Our aim in this chapter is to review the basic concepts and algorithms explored so far, and to look ahead to future developments. We begin in the next section with some fundamental concepts needed to handle heterogeneity in latency, and then provide a broader categorisation of other types of heterogeneity not yet explored in detail. The remainder of the chapter builds a deeper understanding of how heterogeneity has been handled so far, and the prospects for further work.


\section{Fundamental concepts and types of heterogeneity}\label{sec:FundConcepts}

Handling heterogeneous functions relies first on the usual definitions used in multiobjective optimization (available elsewhere in this book). Added to that, we need some well-defined notion of a time budget if we are going to handle problems with different latencies across the objectives. Let's review the required definitions.

\subsection{Fixed evaluation budget definitions}\label{definitions}
As in some of our previous work~\cite{allmendinger2013hang,ejor:Latencies}, we adopt the notion of a fixed evaluation budget in our optimization~(cf.\cite{jansen2012fixed}), as this framework is central to the practical problem of handling heterogeneity in evaluation costs (specifically latency differences between objective components). 

\begin{definition}{(Total budget)}
The total budget for solving an optimization problem is the total number of time steps $B$ available for solving it, under the assumption that only solution evaluations consume any time.\label{TotBudget}
\end{definition}

\begin{definition}{(Limited-capacity parallel evaluation model)}
We assume parallelization of the evaluation of solutions is available, in two senses. First, a solution may (but need not) be evaluated on one objective in parallel to its being evaluated on another objective. Secondly, a number of (at most $\lambda$) solutions may be evaluated at the same time (i.e., as a batch or population) on any objective, provided their evaluation is started at the same time step, and finishes at the same time step (i.e., batches cannot be interrupted, added to, etc., during evaluation). For sake of simplicity, we assume $\lambda$ is the same for all objectives.\label{CapEvalMode}
\end{definition}

\begin{definition}{(Per-objective latency)}
Assume that each objective $i$ can be evaluated in $k_i \in \mathbb{Z}^+$ time steps (for a whole batch). Here, we consider a bi-objective case, and for simplicity, we define $k_1=1$ and $k_2=\ks>1$, so that the slower objective is $\ks$ times slower than the faster one.\label{PerObLatency}
\end{definition}

\begin{definition}{(Per-objective budgets)}
From Definitions~\ref{TotBudget}-~\ref{PerObLatency}, it follows that the total budget of evaluations per objective is different. The budget for $f_1$ 
is $\lambda B$,
 whereas the budget for $f_2$
is $\lambda \lfloor B/\ks \rfloor$. 
In Algorithm~\ref{alg:Heterogeneous} (Section~\ref{algorithms}), we will be referring to the fast and slow objective as $f^{\text{fast}}$ and $f^{\text{slow}}$, respectively, and to their per-objective budgets as $\mathit{MaxFE}^{\text{fast}}$ and $\mathit{MaxFE}^{\text{slow}}$. 
\end{definition}

\subsection{Types of heterogeneity}

We believe that heterogeneous objectives are the norm in multiobjective optimization rather than the exception. Nevertheless, it is still a largely unexplored topic to understand how each different type of heterogeneity causes specific difficulties to existing multiobjective techniques. In many cases, e.g., in industry, the heterogeneity in objectives is perhaps just handled in an {\em ad hoc}\/ way, with some adaptations to existing algorithms. Whether or not such {\em ad hoc}\/ solutions are effective remains an unanswered question for which there is a lot of scope for academic or more foundational work.

Heterogeneity in the latency of different objectives is perhaps the area where it is obvious that the usual {\em ad hoc} solution (``waiting'' for all objectives of a solution to be evaluated) is in need of re-thinking, as our work has shown (for two objectives only so far --- but see the remainder of this chapter for a sketch of a generalization of this).

More generally though, the different types of heterogeneity that might need accounting for in many-objective algorithm design, are as follows (based on but extending~\cite{eichfelder4}):
\begin{enumerate}[i]
    \item Scaling: Different ranges of objective function values are handled by most EMO approaches using either Pareto ranking or dynamic normalization techniques.
    \item Landscape: Variation in landscape features, such as ruggedness, presence of plateaus, separability, or smoothness. Exploratory landscape analysis (ELA)~\cite{MerBisTraPreuWeiRud11:gecco} can help to characterize the landscapes associated with the individual objective functions but there is a lack of research on characterizing the complexity of a multi-objective optimization (MO) problem as a whole (e.g., front shape, local fronts).
    \item Parallelization: Batch vs Sequential evaluation requirements may vary across objectives, and this would complicate the control flow of most types of optimization algorithms.
    \item White vs Grey vs Black Box objectives: A mix of these across the objectives in a single problem would necessitate coordination of different types of searching behaviour, e.g., for Black Box, evolutionary techniques might be effective, but for white box problems, a more efficient search based on knowledge of the objective function may be possible.
    \item Subject to interruptions~\cite{OrseauArmstrong2016,allmendinger2014tuning} or ephemeral resource constraints (ERCs)~\cite{allmendinger2013handling}: If evaluating an objective value of a solution depends upon an experiment which depends upon a resource (such as availability of equipment), then for certain solutions, it may be interrupted if the resource becomes unavailable. This is already an involved problem in the single objective case (ibid).
    \item Variable type (integer, continuous or mixed): These different types usually require different types of algorithm which would need coordination in a multiobjective one.
    \item Determinism: In stochastic settings, objective functions values may depend upon hidden uncontrolled variables necessitating some way of accounting for this variation such as robust or distributionally robust approaches~\cite{bertsimas2019adaptive}.
    \item Noisy vs noiseless: The usual way of handling output (e.g. measurement) noise on objective values is to re-evaluate them to obtain a mean; it would be wasteful to re-evaluate deterministic objectives however --- this seems easily handled, but perhaps there is hidden difficulty.
    \item Theoretical and practical difficulty: This relates to the difficulty of finding the optimum as a whole, not the cost or complexity of evaluating. Combining a simpler objective with a harder one might cause EMO methods to be biased.
    \item Safety~\cite{AllKno2011ecta,KimAllLop2020safe}: This relates to safe optimization, and the tightness of an objective's safety threshold. Evaluating a (non-safe) solution that has an objective value below the safety threshold causes an irrecoverable loss (e.g., breakage of a machine or equipment, or life threat).
    \item Correlations between objectives: Conflicting objectives are considered the norm in multiobjective optimization. However, anti-correlated objectives in particular can lead to very large Pareto fronts, and hence difficulties for multiobjective algorithms (see~\cite{KnoCor2003emo,PaqStu06:mqap}). This may be exacerbated in cases of \emph{many}\/ conflicting objectives, and algorithm settings would need to be carefully chosen to handle these situations~\cite{PurFle2007tec}.  
     \item Evaluation times (latency): This is the topic explored most fully to date, particularly in~\cite{allmendinger2013hang,ejor:Latencies,chugh2018surrogate,10.1145/3377930.3390147,thomann2019representation,thomann2019trust,thomann2019trust_PhD}.
\end{enumerate}
Of course, several of these heterogeneities may exist together in a single problem, which would usually make things even more challenging to handle. Having said that, under certain circumstances, heterogeneity can improve performance. For example, in~\cite{ejor:Latencies,chugh2018surrogate} we observed that a low level of latency between the fast and slow objective can lead to improved results and help reach parts of the Pareto front that may not have been reached otherwise.


\section{Algorithms and benchmarking}\label{sec:algorithms}
In this section we will describe several existing algorithm schemes and benchmark problems/considerations when tackling MO problems with heterogeneous objectives, in particular differing latencies. For this we will refer back to the definitions provided in Section~\ref{definitions}.

\subsection{Algorithms}\label{algorithms}
\begin{figure}[tb!]
    \centering
    \includegraphics[width=1.22\textwidth]{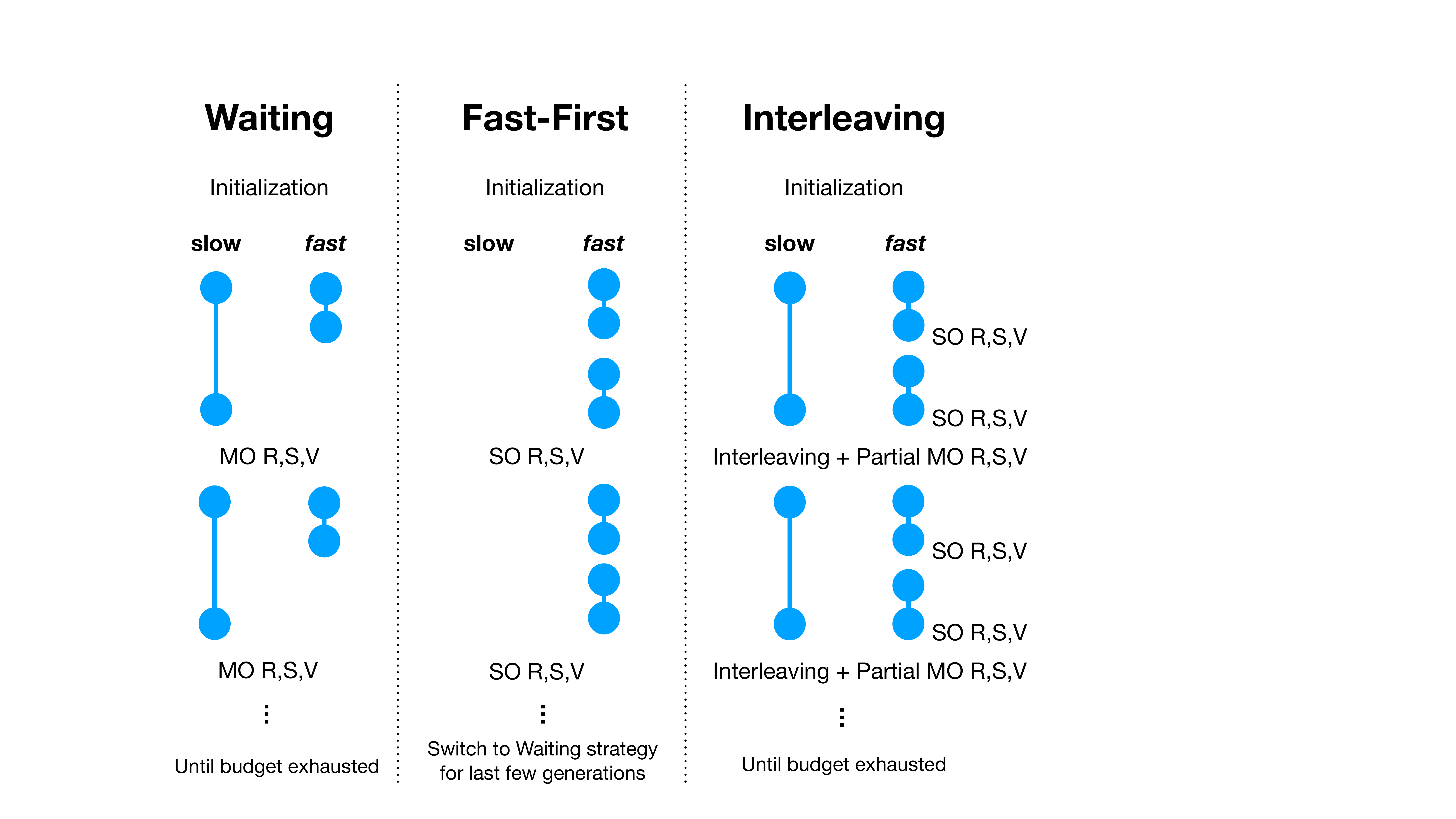}
    \caption{Schematic of the main types of strategy for handling heterogeneous latencies. A biobjective problem is assumed with a slower and a faster objective function (the faster objective is twice as fast as the slower one in the figure, without loss of generality). Further, it is assumed that we are interested in performing well (in a multiobjective sense) given a fixed budget of total evaluation time, and a limited capacity parallel evaluation model (see Section~\ref{definitions}). Three different general strategies are shown --- Waiting, Fast-First, and Interleaving. The time axis is from top to bottom. After initialization, individual solutions need to be evaluated in a parallel batch (population), before entering the usual evolutionary algorithm phases of ranking, selection and variation (R,S,V). When these phases are carried out using evaluations from both objectives, we denote it as MO R,S,V. When only one objective has been evaluated, we denote it as SO R,S,V. In Waiting, only MO R,S,V is used (and a standard MOEA can be employed). In the Fast-First strategy, SO R,S,V is used for some generations, and then subsequently Waiting is used for the remaining generations. In Interleaving, a much more complicated approach, SO R,S,V and MO R,S,V are both used, and solutions evaluated partially and fully are interleaved so that there is less `dead time' than in Waiting, and more guidance than in Fast-First. Interleaving is generally the best. Fuller algorithmic details of these strategies are given in pseudocode in the original papers~\cite{allmendinger2013hang,ejor:Latencies}.}
    \label{fig:schematic}
\end{figure}

This section outlines the three existing algorithm schemes for coping with differing latencies, Waiting, Fast-First, and Interleaving. Figure~\ref{fig:schematic} provides a schematic of these three schemes. 
\\

\noindent \textbf{Waiting. \;} The most straightforward strategy to deal with varying latencies of objectives is to go at the rate of the slow objective, thus fully exploiting the per-objective budget of the slow objective and only partially of the fast objective. This approach avoids the development of customised strategies and is applicable to many-objective problems. In~\cite{ejor:Latencies}, this was referred to as a \textit{Waiting strategy}. 
It prevents the introduction of search bias, for example, towards the fast objectives, and it has shown to perform better as the evaluation budget increases. 
\\

\noindent\textbf{Fast-First. \;}
Strategies falling into this category neglect the time-consuming objective function(s) as long as possible in order to make potentially better use of the time budget by performing a search directed by the cheaper objective function(s). The approach was first proposed in~\cite{ejor:Latencies}, where it was called the Fast-First strategy. It evaluates solutions at the rate of the fastest objective (ignoring other objectives) using a standard (single-objective) EA for part of the optimization, and then switches late (as late as seems reasonable) in the optimization run to a final, evaluation of some selected solutions on the other (slower) objectives to ensure that at least some solutions are evaluated on all objectives. That is, this approaches uses fully the per-objective budget of the fast objective, but only a fraction of the per-objective budget of the slow objective(s). 
Fast-First has shown to perform well for problems with highly positively correlated objectives (for obvious reasons), being able to reach parts of the Pareto front (extreme solutions on the fast objective) that may not have been reached otherwise. Moreover, Fast-First is also almost unaffected by the length of latency, and performed better within a generational multiobjective EA (MOEA) than within a steady state-based MOEA~\cite{ejor:Latencies}. Similar to Waiting, Fast-First is readily applicable to many-objective problems. 
\\

\noindent\textbf{Interleaving. \;} These strategies are less straightforward since they employ a mechanism to coordinate the evaluation of the objectives during search so as to use the per-objective budgets of all objectives as efficiently and thoroughly as possible. Also, so far Interleaving strategies have been applied to bi-objective problems (one slow and one fast objective) only.

In our initial work on latencies~\cite{allmendinger2013hang}, we proposed an Interleaving strategy embodied by a \textit{ranking-based EMOA} that maintained a population unbounded in size, and which assigned pseudovalues to a solution's slow objective until that objective has been evaluated. Different techniques to assign pseudovalues have been proposed including one based on fitness-inheritance. 
The approach works similarly to a standard ranking-based EMOA, where offspring are generated by a process of (multiobjective) selection, crossover and mutation. Every time a batch of solutions has been evaluated on the slow objective, their pseudovalues are replaced with the true objective values, and a new batch of solutions (selected from the current unbounded population) is submitted for evaluation on the slow objective, and new pseudovalues assigned to these solutions' slow objectives. This new batch consists either of the most recently generated solutions that have not been evaluated on the slow objective yet, or of solutions selected based on their anticipated quality computed based on their non-dominated sorting rank. 
This very first approach to cope with heterogeneous evaluation times performed well for long latencies when compared to a Waiting scheme. 

In~\cite{ejor:Latencies}, we proposed two further variations of interleaving strategies, \textit{Brood Interleaving (BI)} and \textit{Speculative Interleaving (SI)}. As in the approach explained above, both strategies evaluate solutions on both objectives in parallel. However, BI and SI employ a constant population size, and use the time while the slow objective is being evaluated (the interleaving period) to generate and evaluate solutions on the fast objective only: The Brood Interleaving strategy generates these solutions using uniform selection and variation applied to the population currently evaluated on the slow objective, while SI initializes an inner (single-objective) EA with this population and applies it to the optimization of the fast objective for the remainder of the interleaving period. The solutions evaluated on the fast objective are then used as a quality indicator to decide which solutions to evaluate on the slow objective in the next generation.\footnote{Solutions evaluated on the fast objective are considered for evaluation on the slow objective in the next generation if they outperform at least one of their parents on the fast objective.}
The difference between SI and BI translates into deliberately optimizing the fast objective vs maintaining selection pressure where possible. As shown in~\cite{ejor:Latencies}, the Speculative Interleaving strategy performs well for low evaluation budgets and/or when latencies are long, objectives positively correlated, and fitness landscapes rugged. Similar to Waiting, BI performs well for larger evaluation budgets, with BI performing better for longer latencies and larger search spaces. Furthermore, BI performs significantly better when used in combination with a steady state-based MOEA than a generational-based MOEA. 


\begin{algorithm}[t]
\KwIn {$\mathit{MaxFE}^{\text{slow}}$ and $\mathit{MaxFE}^{\text{fast}}$: per-objective budget of slow ($f^{\text{slow}}$) and fast objective ($f^{\text{fast}}$); $\ks$: latency; $\lambda$: initial population size; $u$: number of new samples per iteration; $\tau$: transfer learning trigger}
\KwOut{Non-dominated solutions of the archive $A$}
Create an initial population \textit{P}, set archives to $A=A^{\text{fast}}:=\emptyset$, iteration counter to $i:=0$, and evaluation counters to $\mathit{FE}^{\text{slow}}=\mathit{FE}^{\text{fast}}:=0$\\
\While{$P$ $\text{\upshape is evaluated on}$ $f^{\text{slow}}$}{
Evaluate $P$ on $f^{\text{fast}}$, and add solutions to $A^{\text{fast}}$ \\
Run a single-objective EA to optimize $f^{\text{fast}}$ using $\lambda \times (\ks- 1)$ function evaluations, and add the solutions to $A^{\text{fast}}$ 
}
Update archive and counters to $A:=P$, $\mathit{FE}^{\text{slow}}:=\lambda$, $\mathit{FE}^{\text{fast}}:= \lambda\times\ks$, $i:=i+1$\\
\While{$\mathit{FE}^{\text{slow}}$ < $\mathit{MaxFE}^{\text{slow}}$ $\text{\upshape and}$ $\mathit{FE}^{\text{fast}}$ < $\mathit{MaxFE}^{\text{fast}}$}{
\uIf {HK-RVEA}{
Build surrogates for the slow and fast objective function based on $A$ and $A^{\text{fast}}$\\
Run a multiobjective EA (RVEA~\cite{cheng2016reference}) to find samples for updating the surrogates\\
Form new population $P$ by selecting $u$ samples using the acquisition function from the K-RVEA algorithm~\cite{chugh2016surrogate}\\
}
\uIf {T-SAEA}{
Build surrogates for the slow and fast objective function based on $A$ and $A^{\text{fast}}$, but, every $\tau$ iterations (if $i\mod{\tau}=0$), build the surrogate of the slow objective function using a transfer learning approach\\
Run a multiobjective EA (RVEA~\cite{cheng2016reference}) to find samples for updating the surrogates\\
Form new population $P$ by selecting $u$ samples using an adaptive acquisition function~\cite{wang2020adaptive} followed by the angle-penalized distance approach (taken from RVEA~\cite{cheng2016reference})\\
}
\While{$P$ $\text{\upshape is evaluated on}$ $f^{\text{slow}}$}{
Evaluate $P$ on $f^{\text{fast}}$, and add solutions to $A^{\text{fast}}$ \\
\uIf {HK-RVEA}{
Create $u\times(\ks- 1)$ solutions via uniform selection and variation applied to $P$\\}
\uIf {T-SAEA}{
Create $u\times(\ks- 1)$ solutions via Latin hypercube sampling around $P$\\
}
Evaluate newly created solutions on $f^{\text{fast}}$, and add them to $A^{\text{fast}}$
}
Update archive and counters to $A:=A\cup P$, $\mathit{FE}^{\text{slow}}:=\mathit{FE}^{\text{slow}}+u$, $\mathit{FE}^{\text{fast}}:= u\times\ks$, $i:=i+1$\\
}
\caption{{Interleaving surrogate-based strategies (HK-RVEA, T-SAEA)}
\label{alg:Heterogeneous}}
\end{algorithm}

Two surrogate-based interleaving strategies for coping with latencies in the objectives (one fast and one slow) have been proposed recently in~\cite{chugh2018surrogate} (HK-RVEA) and ~\cite{10.1145/3377930.3390147} (T-SAEA); see Algorithm \ref{alg:Heterogeneous} for a sketch of the two algorithms. Although these and the non-surrogate-based strategies outlined above adopt the same limited-capacity parallel evaluation model, the surrogate-based methods opted to use a different value of $\lambda$ (number of solutions evaluated in parallel) during search: a large $\lambda$ is used to create the initial training data set, while a significantly smaller $\lambda$ is used thereafter. Identifying and evaluating fewer samples in each iteration, but doing more iterations, is more suitable for a surrogate-based approach; in fact, traditional surrogate-based methods, e.g. see~\cite{JonSchWel98go}, use one sample per iteration. Consequently, in~\cite{chugh2018surrogate,10.1145/3377930.3390147}, the stopping criterion was not the maximum number of time steps (or iterations), as used by the non-surrogate-based methods, but the maximum number of per-objective function evaluations only (ignoring the number of time steps used). In practice, the stopping criteria (times steps vs function evaluations only) are dictated by the problem (context) at hand, thus making certain surrogate-based methods potentially unsuitable. For example, in some of our closed-loop optimization work~\cite{Kno2009closed}, the experimental platform dictates that a batch of solutions of a certain size should be evaluated in parallel (a 96-well plate was used in some experiments, in others a microarray for assaying 9,000 DNA strands was available).

The strategy proposed in~\cite{chugh2018surrogate}, called \textit{HK-RVEA}, resembles a hybrid between BI and SI combined with a surrogate-assisted approach for selecting solutions to be evaluated on the slow and fast objective. Following initialization of the population and submitting it for evaluation on both objectives, HK-RVEA uses a single-objective EA (without any surrogate) to optimize the fast objective (like in SI) whilst the slow objective is being evaluated. In the main loop, the EA is replaced by repetitively applying crossover and mutation to the population (as in BI) as the number of solutions (samples) evaluated on the slow objective is much lower than the initial population size. 
HK-RVEA maintains a separate archive of evaluated solutions for the slow and fast objective, and then uses these archives (i.e., different samples) to build an objective-specific surrogate. A multiobjective EA (RVEA~\cite{cheng2016reference}) is used to find (three) samples for updating the surrogates using the infill criteria from the K-RVEA algorithm~\cite{chugh2016surrogate}. HK-RVEA has been shown to perform well for problems with short latencies, occasionally even outperforming a multiobjective EA (K-RVEA) optimizing the same problem but without latency.


The approach proposed in~\cite{10.1145/3377930.3390147} is called \textit{T-SAEA} and it varies from HK-RVEA primarily in that it combines a surrogate-assisted evolutionary algorithm with a transfer learning approach, which is used to update the surrogate for the slow objective (for most of the time). The motivation is that if there is a strong similarity or correlation between the slow and fast objective, then knowledge transfer is beneficial, otherwise negative transfer may occur. The basis of the proposed transfer learning approach is a preceding filter-based feature selection method~\cite{cervante2012binary} adopted to identify the most relevant decision variables to share between the surrogate of the slow and fast objective. Based on the identified subset of decision variables, an adaptive aggregation method is then used to share the parameters of the two surrogates. The adaptive component, in essence, shifts the importance of sharing parameters of the fast objective surrogate to the parameters of the slow objective surrogate as the optimization progresses. T-SAEA has shown significantly better performance~\cite{10.1145/3377930.3390147} compared to non-surrogate methods designed for handling latencies. The algorithm seems to do well also against HK-RVEA, 
allowing us to tentatively conclude that transfer learning is a promising approach to cope with latencies, provided the non-trivial issue around negative transfer (decrease in learning performance in the target domain) can be addressed.
\\

\noindent\textbf{Non-evolutionary approaches. \;}The approaches discussed above employ evolutionary search at some stage. Thomann and co-authors~\cite{thomann2019trust,thomann2019trust_PhD} propose a non-evolutionary approach based on the trust region method, which optimizes one point at a time. The method is called \textit{MHT} (short for multiobjective heterogeneous trust region algorithm) and it differs from other trust region methods in the way the search direction is computed and the replacement of the objectives by quick-to-evaluate surrogates. In fact, MHT replaces all objectives (slow and fast ones) with a local quadratic model that interpolates the current iteration point, and is agnostic about the per-objective latencies. The next iteration point is determined by first solving the classical trust region subproblem to obtain the ideal point, followed by solving an auxiliary problem (know as the Tammer-Weidner functional~\cite{gerth1990nonconvex}) to determine a trial point. The trial point is accepted as the next iteration point if a multiobjective condition describing the improvement of the function values is fulfilled. Otherwise, the current point is kept and the size of the trust region (the trust region radius) reduced. MHT stops when there is no improvement in the iteration point. To keep the number of evaluations of the slow objective to a minimum, MHT evaluates the slow objective only when the surrogate becomes inaccurate.

MHT is scalable to any number of objectives and any combination of fast vs slow objectives (while the approaches above assumed a bi-objective problem with one fast and one slow objective, though Waiting and Fast-First are easily scalable). However, the assumption taken here is that while the expensive functions are black box (and slow to compute), the fast objectives are given as analytical functions for which function values and derivatives can easily be computed (above we assumed that the fast objective can be black box and there is a budget on how often the fast objective can be evaluated). Moreover, it is assumed that the slow (black-box) objectives are twice continuously differentiable, which is a strong assumption to make, as noted by the authors~\cite{thomann2019trust_PhD}.

In~\cite{thomann2019representation,thomann2019trust_PhD}, Thomann and Eichfelder propose three heuristics augmented onto the standard version of MHT outlined above. The purpose of these heuristics, which are motivated by ideas for bi-objective optimization problems, is to exploit the heterogeneity of the objective functions further to identify additional Pareto optimal solutions that are spread over the Pareto front. For the heuristics to be applicable it is assumed that the fast objective can be optimized with reasonable numerical effort, and that it is bounded from below (assuming a minimization problem); in these heuristics the fast objective is not replaced by a surrogate. Below we explain these heuristics briefly.

The idea of the first heuristic, referred to as \textit{Spreading}, is to minimize the fast objective on local areas that move in the direction of the optima of the fast objective. A local area can be seen as a trust region (defined by user-provided radius, or spreading distance) around an optimal input point. The initial (weakly) optimal input point can be obtained, e.g., by applying the standard version of MHT. The point in this local area that minimizes the fast objective only, can then be used as the starting point for the next run of MHT. Repeating this process allows one to successively identify new optimal (weakly efficient) input points until the global optimal value of the fast objective has been reached. Finally, dominated solutions are deleted to leave optimal solutions only. Note, the closer the initial optimal point to the global minimal value of the fast objective, the fewer successive optimal points can be computed. Also, the larger the spreading distance, the bigger the distance between the computed points and thus the fewer optimal points can be obtained.

The second heuristic, \textit{Image Space Split}, splits the objective space into disjoint areas (which can be thought of as slices in a two-dimensional objective space) in which then a modified version of MHT is applied. The reason for needing a modified version of MHT to accommodate this heuristic is that the presence of a lower bound on the values of the fast objective requires an additional constraint for computing the ideal point and a modified auxiliary problem for determining the descent direction; all other steps in MHT remain unchanged. When splitting the objective space into several disjoint search areas before applying any version of MHT, the choice of starting points in the disjoint search areas is important. A heuristic approach to determine starting points is suggested, and so is a heuristic stopping criterion to save function evaluations. The challenge for this heuristic is to decide on a suitable number of disjoint search regions. In general, the greater this number, the more efficient points are computed and the more function evaluations are required. However, this does not need to be always the case because of the non-linear relationship between the location of the individual search regions and the position of the starting points, resulting in scenarios where not all search regions contain (weakly) optimal points and some starting points being already close to efficient points.

The third heuristic is a \textit{Combination of Image Space Split and Spreading}. This heuristic first executes Image Space Split, and then applies Spreading in the space captured by both adjacent optimal solutions identified through Image Space Split to compute further optimal points. The challenges associated with the two individual heuristics (merged together) persist in this heuristic. 


An initial study~\cite{THOMANN2019104103}, comparing MHT against a weighted sum-based approach and direct multisearch for multiobjective optimization, concluded that the proposed approach can yield good results. 

\subsection{Empirical study: Towards many-objective heterogeneous latencies}
In this section we will present and analyze original empirical results concerning the relationship between the number of objectives and the level of heterogeneity one might expect in a problem; all this is done in the context of varying latencies. 

\begin{figure}[!t]
\centering
\includegraphics[height=170pt]{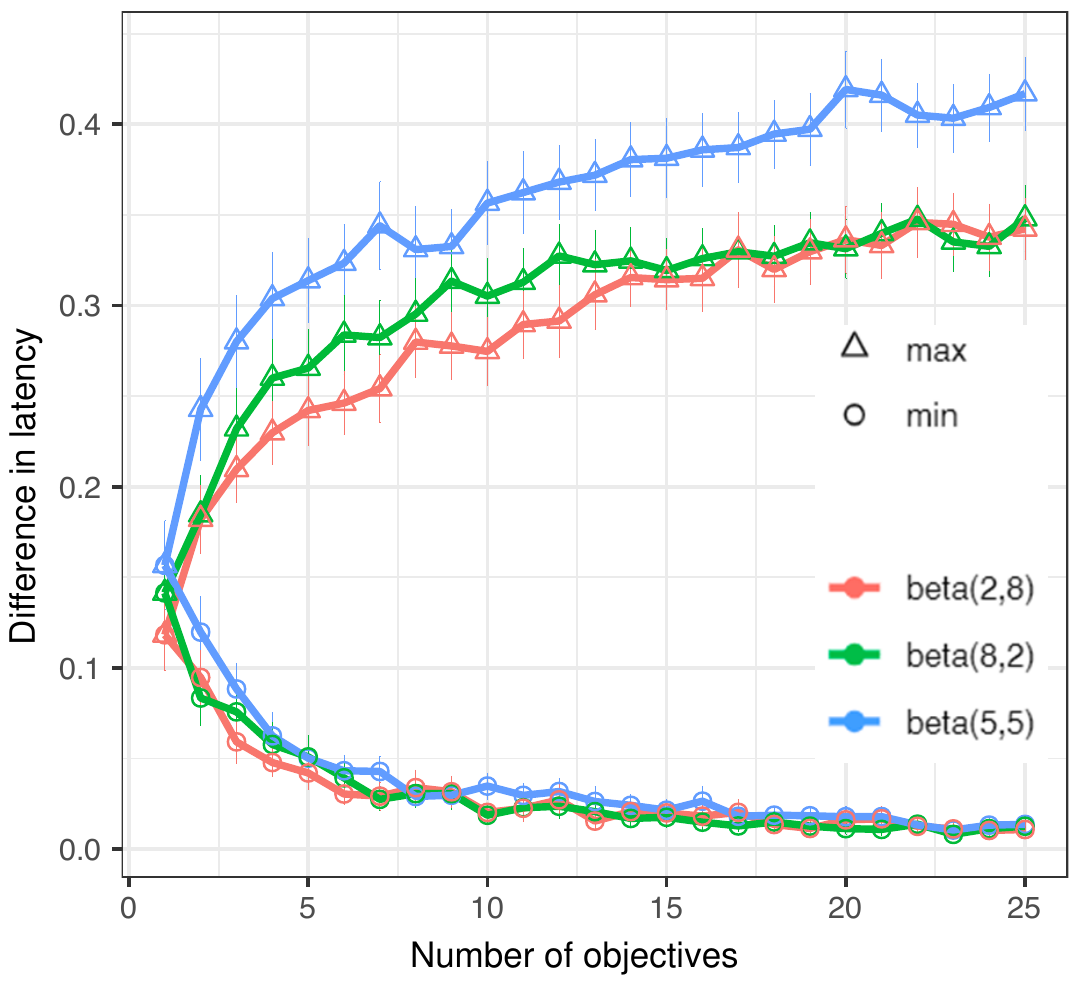}
\caption{Mean and standard error of the maximum (triangles) and minimum differences (circles) in per-objective latencies (y-axis) as a function of the number of objectives in a problem (x-axis). Per-objective latencies are drawn from one of three Beta distributions.}
\label{fig:HetOb}
\end{figure}

For this experiment, assume a problem with a certain number of objectives, each being associated with a per-objective latency drawn from a given distribution. We want to understand the likely range of latencies among the objectives. Having a better understanding about the level of heterogeneity can help us in the design and selection of suitable algorithms. The experiment consists of creating problems that vary in the number of objectives (1-25 objectives) with per-objective latencies drawn from one of three Beta distributions --- $\texttt{beta}(2,8)$ (skewed to the right), $\texttt{beta}(8,2)$ (skewed to the left), and $\texttt{beta}(5,5)$ (symmetric), each defined on the interval [0,1]. Each combination of objective number and distribution ($25\times3=75$ combinations in total), was realized 100 times, and the mean and standard error of the minimum and maximum differences in per-objective latencies plotted in Figure~\ref{fig:HetOb}. The reason to use a Beta distribution is that it allows for a convenient simulation of skewness in the per-objective latencies, and due to its wide-spread use in the literature and practice to quantify the duration of tasks (see seminal paper of~\cite{PERT1959}).

Two main observations can be made from the figure:
\begin{itemize}
    \item Regardless of the probability distribution (skewness) used, the mean of the minimum and maximum difference in per-objective latencies starts roughly from the same level for a bi-objective problem, with the mean differences then increasing/decreasing logarithmically with the number of objectives. This pattern is due to sampling from a wide distribution, where it becomes gradually more likely that the per-objective of a new objective is either more similar or more distinct to the per-objective latencies of an existing objective. If all per-objective latencies were identical, then the minimum and maximium difference in per-objective latencies would be identical and of value 0 for all objectives. 

    \item There is an asymmetry between the minimum and maximum difference in the per-objective latencies as a function of the number of objectives. The mean minimum difference is very similar for the three Beta distributions with the mean difference flattening quickly beyond around 15 objectives. However, there is a statistical difference between the Beta distributions when considering the mean of the maximum differences. In particular, the mean maximum difference obtained with the symmetric distribution ($\texttt{beta}(5,5)$) increases more rapidly than with the mean maximum difference of the two skewed Beta distributions. This pattern is due to the higher probability of sampling extreme per-objective latencies with the symmetric Beta distribution. 
\end{itemize}

The conclusion of this experiment is that, given knowledge of how many objectives a problem has, and even a limited knowledge about the distribution of per-objective latencies, then one can estimate with reasonable accuracy the level of heterogeneity in terms of the greatest latency difference one will observe in practice. This should facilitate the detailed design of algorithms one might consider for handling the
heterogeneity.


\subsection{Benchmarking}
Validating algorithms designed for coping with heterogeneity in the objective evaluation times requires careful consideration of how evaluations are counted/simulated, what test problems to use, and how to validate performance. Below we will discuss how these are affected in an environment with heterogeneous objectives. 

\subsubsection{How to count/simulate evaluations?}
In a standard multi/many-objective optimization problem, where homogeneous evaluation times of objectives are assumed, one evaluation is typically equivalent to evaluating one solution on all objectives that the problem has with the stopping criterion being a given maximum number of evaluations. When simulating a real problem with latencies, the maximum number of function evaluations available can be different for each of the objectives, meaning not all solutions can be evaluated on all objectives. Moreover, depending on the problem at hand, the stopping criteria may indeed be only the maximum number of function evaluations on both objectives (regardless of the number of time steps used up) (as used in~\cite{chugh2018surrogate,10.1145/3377930.3390147}) but it can also be the maximum number of time steps available (as used in~\cite{ejor:Latencies,allmendinger2013hang}). The latter would apply in the case where the optimization process has to be terminated within a certain time frame. 

When simulating a real problem with latencies, it is also important to reveal objective values to the optimizer only when the simulated evaluation of an objective is complete. If solutions can be evaluated in a batch, then multiple objective values are revealed in one go.



In~\cite{thomann2019trust,thomann2019representation,thomann2019trust_PhD}, the difference in evaluation times of the slow and fast objective was not simulated, solutions were not evaluated in parallel, and search was terminated if there is no improvement in a solution's objective function values. Also, since the slow objective was evaluated only if the surrogate model was not accurate enough, it is unknown a priori how often the slow and fast objective will be evaluated. This setup needs to be taken into account when tackling a practical problem.

\subsubsection{Test problems}
In an ideal world, one would evaluate an algorithm on a real problem featuring heterogeneous evaluation times. However, this is not feasible because it would usually be too cost and time prohibitive, especially considering that a (stochastic) algorithm would need to be executed multiple times on a problem to obtain information about statistical significance in performance results. This is a typical issue in expensive optimization.

Consequently, it is suggested and also accepted by the community, to adapt existing test problems to simulate heterogeneity in the objectives. This is the approach taken to validate all algorithms described above. To simulate heterogeneity in the objective evaluation time, one can simply take any existing multi/many-objective test problem, and declare one/some of the objectives as slow (expensive) and also specify by how much these are slower than the other objectives. In principle, this way one can create a problem where all objectives differ in their evaluation time. Consequently, one can study the impact on algorithm performance of different ratios between the slow and fast objective(s), as done, for example, in~\cite{allmendinger2013hang,ejor:Latencies,chugh2018surrogate,10.1145/3377930.3390147}. Of course, if the underlying problem has configurable problem parameters, then the impact of these on heterogeneous evaluation times can be investigated too. For example, in~\cite{ejor:Latencies,chugh2018surrogate} we proposed a binary and continuous bi-objective toy problem with configurable correlation levels between the fast and slow objective, in~\cite{ejor:Latencies} we investigated the impact of the landscape ruggedness on algorithm performance (using MNK landscapes), and in~\cite{thomann2019trust_PhD} the impact of varying the number of decision variables was investigated. 

All research so far on heterogeneous objective evaluation times considered problems with exactly one slow objective, and at least one fast objective. Typically, a random objective or the most difficult objective (as done in~\cite{thomann2019trust_PhD}) was declared as the slow objective.

\subsubsection{Validating performance}
It is obvious that any new proposed algorithm for dealing with heterogeneous objective evaluation times should be compared against other methods designed for the same purpose. As a baseline, it makes sense to compare with Waiting (to mimic a naive approach), and, to upper bound performance, to optimize the same problem without heterogeneity (i.e., assume that all objectives have the same evaluation time).
 
So far, standard performance metrics designed for multiobjective optimization were used to measure algorithm performance, such as IGD and the hypervolume metric, and, for the non-evolutionary approaches proposed in~\cite{thomann2019trust_PhD}, the number of Pareto optimal solutions and functions evaluations needed to discover these were recorded. To understand visually whether the heterogeneity in evaluation times introduced any search bias (e.g., towards the optimization of the fast objective), plotting the median attainment surface is a reasonable approach (though this is applicable to bi and tri-objective problems only). 

\section{Related research}\label{sec:RelatedResearch}

Research on heterogeneous objectives can be related to and gain inspiration from a number of other areas in the literature. In the following we briefly discuss some of these relationships. 

An obvious connection exists with the use of asynchronous evolutionary algorithms in MO in distributed environments as arising, for example, in grid computing, multi-core CPUs, clusters of CPUs or on virtual clouds of CPUs~\cite{scott2015evaluation,yagoubi2011asynchronous,lewis2009asynchronous}. Here it is assumed that the cloud computing resource induces heterogeneity and/or unreliability causing the evaluation time of a solution to vary across computing resources. This introduces asynchronicity across the population and not across the individual objectives, which is our focus. Despite considering a different problem setup, the research questions in the two areas are similar and are centred around understanding (i)~how to utilize efficiently the available resources to perform MO in the presence of heterogeneous resources (per solution vs per objective) and (ii)~the bias induced by heterogeneity on the search path taken by the optimizer (bias towards search regions quick to evaluate solutions vs quick to evaluate objectives
). 

Machine learning methods focused on dealing with missing data~\cite{little2019statistical,garcia2010pattern,allison2001missing} and surrogate models~\cite{allmendinger2017surrogate,tabatabaei2015survey,SnoLarAda2012nips} are of importance when designing algorithms for coping with heterogeneous objectives. In particular, these methods can be used to substitute missing objective function values with proxies and substitute expensive functions with an approximation function that is cheap to evaluate, respectively. This is relevant, for example, when dealing with heterogeneity in the evaluation times of objectives, and MO problems that are subject to interruptions or ERCs. The application of surrogate-assisted methods to batch optimization~\cite{azimi2010batch,gonzalez2016batch,chugh2016surrogate} and asynchronous batch optimization~\cite{ginsbourger2011dealing,SnoLarAda2012nips} is also highly relevant to multiobjective problems with differing latencies. 

The application of transfer learning~\cite{pan2009survey}, potentially combined with dimensionality reduction~\cite{van2009dimensionality}, as done by T-SAEA~\cite{10.1145/3377930.3390147} (see above), is another machine learning methodology that can find application to problems with heterogeneous objectives beyond problems with latencies. Methods for dimensionality reduction can also be applied to the objective space~\cite{brockhoff2009objective} with the aim of homogenizing the objectives. For example, removing an expensive objective that is positively correlated with a cheap objective, would reduce the total number of objectives and homogenize the set of remaining objectives, simplifying their optimization. A potential issue with this approach is that the level of correlation between objectives is a statistical observable that may not hold all over the entire search space and in particular may not hold crucially at or near optimality. 

Scheduling concepts and methods~\cite{gonzalez2016batch,kalashnikov2013mathematical,talbi2008parallel,brucker1999resource} are also of relevance for MO problems with heterogeneous objectives, especially when faced with latencies or ERCs. Here, scheduling methods can help, for example, to decide on when to evaluate which solution on which objective such that, for example, idle time where no objective is being evaluated is minimized, and existing resources required for evaluating an objective are being utilized most efficiently. To cope with heterogeneous objectives, inspiration can be gained, for example, from classical resource-constrained scheduling~\cite{brucker1999resource}, parallelization of MO algorithms~\cite{talbi2008parallel}, queuing theory~\cite{kalashnikov2013mathematical} and batch expensive optimization~\cite{gonzalez2016batch}.


\section{Conclusions and future work}\label{sec:FutureResearch}

This chapter reviewed the topic of multi/many-objective optimization problems with heterogeneous objectives, meaning the objectives vary in different aspects, such as computational complexity, evaluation effort, or determinism, or a combination of these. Although heterogeneous objectives exist in practical applications, very little research has been carried out by the community to address this problem feature. This chapter started by describing motivational examples of problems with heterogeneous objectives, followed by the introduction of basic concepts and a taxonomy for modelling heterogeneity. We then discussed different types of heterogeneity, described existing algorithms designed for coping with heterogeneous objectives, and reviewed benchmarking considerations arising due to heterogeneity. Finally, we have reviewed related research. The algorithm part of the chapter was focused largely on a particular type of heterogeneity, namely different evaluation times of objectives, as this is the only type that has gained attention in the community, originating from work carried out by the authors of this chapter.

The chapter has highlighted that heterogeneous objectives exist in a range of practical applications, and, although the community has started to look at this issue, there is much more that can and needs to be done. First and foremost, we need to raise awareness in the decision and data science community and amongst practitioners about the meaning of heterogeneous objectives and that we have approaches to cope with this issue. This chapter will hopefully contribute to this aspect. However, at the same time, it needs to be made clear that the existing approaches have limitations, and that the only type of heterogeneity investigated so far is different evaluation times of objectives (where one objective is slow and the others fast to evaluate). More research is needed to extend the existing methods to cope with many-objective problems where the objectives can be of any duration (and not of two modes only, slow vs fast). The development of methods to cope with other types of heterogeneity is needed too, and so are methods for problems where heterogeneity of multiples types exist in one problem. There is also a need for a customized benchmarking process. In particular, configurable many-objective test problems to simulate and adjust heterogeneity are needed, and existing performance metrics and visualisation tools may need to be adjusted/extended to understand further the impact of heterogeneous objectives. We look forward to future progress by the field in these directions.


\bibliographystyle{plain}
\bibliography{References,abbrev,journals,authors,biblio,crossref}

\providecommand{\MaxMinAntSystem}{{$\cal MAX$--$\cal MIN$} {A}nt {S}ystem}
  \providecommand{\Rpackage}[1]{#1} \providecommand{\SoftwarePackage}[1]{#1}
  \providecommand{\proglang}[1]{#1}
\begin{thebibliography}{10}

\bibitem{allison2001missing}
Paul~D Allison.
\newblock {\em Missing data}, volume 136.
\newblock Sage publications, 2001.

\bibitem{Allmendinger2012phd}
Richard Allmendinger.
\newblock {\em Tuning evolutionary search for closed-loop optimization}.
\newblock PhD thesis, The University of Manchester, UK, 2012.

\bibitem{allmendinger2017surrogate}
Richard Allmendinger, Michael~TM Emmerich, Jussi Hakanen, Yaochu Jin, and
  Enrico Rigoni.
\newblock Surrogate-assisted multicriteria optimization: Complexities,
  prospective solutions, and business case.
\newblock {\em Journal of Multi-Criteria Decision Analysis}, 24(1-2):5--24,
  2017.

\bibitem{allmendinger2014tuning}
Richard Allmendinger, Spyridon Gerontas, Nigel~J Titchener-Hooker, and
  Suzanne~S Farid.
\newblock Tuning evolutionary multiobjective optimization for closed-loop
  estimation of chromatographic operating conditions.
\newblock In {\em International Conference on Parallel Problem Solving from
  Nature}, pages 741--750. Springer, 2014.

\bibitem{ejor:Latencies}
Richard Allmendinger, Julia Handl, and Joshua Knowles.
\newblock {Multiobjective optimization: When objectives exhibit non-uniform
  latencies}.
\newblock {\em European Journal of Operational Research}, 243(2):497--513,
  2015.

\bibitem{allmendinger2013hang}
Richard Allmendinger and Joshua Knowles.
\newblock `{Hang} on a minute': Investigations on the effects of delayed
  objective functions in multiobjective optimization.
\newblock In {\em International Conference on Evolutionary Multi-Criterion
  Optimization}, pages 6--20. Springer, 2013.

\bibitem{allmendinger2013handling}
Richard Allmendinger and Joshua Knowles.
\newblock On handling ephemeral resource constraints in evolutionary search.
\newblock {\em Evolutionary computation}, 21(3):497--531, 2013.

\bibitem{AllKno2011ecta}
Richard Allmendinger and Joshua~D. Knowles.
\newblock Evolutionary search in lethal environments.
\newblock In {\em International Conference on Evolutionary Computation Theory
  and Applications}, pages 63--72. SciTePress, 2011.

\bibitem{azimi2010batch}
Javad Azimi, Alan Fern, and Xiaoli~Z Fern.
\newblock Batch {B}ayesian optimization via simulation matching.
\newblock In {\em Advances in Neural Information Processing Systems}, pages
  109--117, 2010.

\bibitem{bertsimas2019adaptive}
Dimitris Bertsimas, Melvyn Sim, and Meilin Zhang.
\newblock Adaptive distributionally robust optimization.
\newblock {\em Management Science}, 65(2):604--618, 2019.

\bibitem{brockhoff2009objective}
Dimo Brockhoff and Eckart Zitzler.
\newblock Objective reduction in evolutionary multiobjective optimization:
  Theory and applications.
\newblock {\em Evolutionary computation}, 17(2):135--166, 2009.

\bibitem{brucker1999resource}
Peter Brucker, Andreas Drexl, Rolf M{\"o}hring, Klaus Neumann, and Erwin Pesch.
\newblock Resource-constrained project scheduling: Notation, classification,
  models, and methods.
\newblock {\em European journal of operational research}, 112(1):3--41, 1999.

\bibitem{cervante2012binary}
Liam Cervante, Bing Xue, Mengjie Zhang, and Lin Shang.
\newblock Binary particle swarm optimisation for feature selection: A filter
  based approach.
\newblock In {\em 2012 IEEE Congress on Evolutionary Computation}, pages 1--8.
  IEEE, 2012.

\bibitem{cheng2016reference}
Ran Cheng, Yaochu Jin, Markus Olhofer, and Bernhard Sendhoff.
\newblock A reference vector guided evolutionary algorithm for many-objective
  optimization.
\newblock {\em IEEE Transactions on Evolutionary Computation}, 20(5):773--791,
  2016.

\bibitem{chugh2018surrogate}
Tinkle Chugh, Richard Allmendinger, Vesa Ojalehto, and Kaisa Miettinen.
\newblock Surrogate-assisted evolutionary biobjective optimization for
  objectives with non-uniform latencies.
\newblock In {\em Proceedings of the Genetic and Evolutionary Computation
  Conference}, pages 609--616, 2018.

\bibitem{chugh2016surrogate}
Tinkle Chugh, Yaochu Jin, Kaisa Miettinen, Jussi Hakanen, and Karthik Sindhya.
\newblock A surrogate-assisted reference vector guided evolutionary algorithm
  for computationally expensive many-objective optimization.
\newblock {\em IEEE Transactions on Evolutionary Computation}, 22(1):129--142,
  2016.

\bibitem{eichfelder4}
Gabriele Eichfelder, Xavier Gandibleux, Martin~Josef Geiger, Johannes Jahn,
  Andrzej Jaszkiewicz, Joshua~D Knowles, Pradyumn~Kumar Shukla, Heike
  Trautmann, and Simon Wessing.
\newblock {Heterogeneous Functions (WG3)}.
\newblock In {\em {Understanding Complexity in Multiobjective Optimization:
  Report from Dagstuhl Seminar 15031 / Greco, Salvatore; Klamroth, Kathrin;
  Knowles, Joshua D.; Rudolph, Günter. --- Wadern : Schloss Dagstuhl. Vol.
  5}}, pages 121--129. Dagstuhl Zentrum f\"ur Informatik, 2015.

\bibitem{garcia2010pattern}
Pedro~J Garc{\'\i}a-Laencina, Jos{\'e}-Luis Sancho-G{\'o}mez, and An{\'\i}bal~R
  Figueiras-Vidal.
\newblock Pattern classification with missing data: A review.
\newblock {\em Neural Computing and Applications}, 19(2):263--282, 2010.

\bibitem{gerth1990nonconvex}
Chr Gerth and P~Weidner.
\newblock Nonconvex separation theorems and some applications in vector
  optimization.
\newblock {\em Journal of Optimization Theory and Applications},
  67(2):297--320, 1990.

\bibitem{ginsbourger2011dealing}
David Ginsbourger, Janis Janusevskis, and Rodolphe Le~Riche.
\newblock {Dealing with asynchronicity in parallel Gaussian Process based
  global optimization}.
\newblock Research report, {Mines Saint-Etienne}, 2011.

\bibitem{gonzalez2016batch}
Javier Gonz{\'a}lez, Zhenwen Dai, Philipp Hennig, and Neil Lawrence.
\newblock Batch {B}ayesian optimization via local penalization.
\newblock In {\em Artificial intelligence and statistics}, pages 648--657,
  2016.

\bibitem{Dagstuhl15031}
Salvatore Greco, Kathrin Klamroth, Joshua~D. Knowles, and G{\"u}nther Rudolph,
  editors.
\newblock {\em Understanding Complexity in Multiobjective Optimization
  (Dagstuhl Seminar 15031)}, volume 5(1) of {\em Dagstuhl Reports}.
\newblock Schloss Dagstuhl--Leibniz-Zentrum f{\"u}r Informatik, Germany, 2015.

\bibitem{jansen2012fixed}
Thomas Jansen and Christine Zarges.
\newblock Fixed budget computations: A different perspective on run time
  analysis.
\newblock In {\em Proceedings of the 14th annual conference on Genetic and
  evolutionary computation}, pages 1325--1332, 2012.

\bibitem{JonSchWel98go}
D.~R. Jones, M.~Schonlau, and W.~J. Welch.
\newblock Efficient global optimization of expensive black-box functions.
\newblock {\em Journal of Global Optimization}, 13(4):455--492, 1998.

\bibitem{kalashnikov2013mathematical}
Vladimir~V Kalashnikov.
\newblock {\em Mathematical methods in queuing theory}, volume 271.
\newblock Springer Science \& Business Media, 2013.

\bibitem{KimAllLop2020safe}
Youngmin Kim, Richard Allmendinger, and Manuel Lopez-Ibanez.
\newblock Safe learning and optimization techniques: Towards a survey of the
  state of the art.
\newblock {\em Arxiv preprint arXiv:2101.09505 [cs.LG]}, 2021.

\bibitem{Kno2009closed}
Joshua~D. Knowles.
\newblock Closed-loop evolutionary multiobjective optimization.
\newblock {\em IEEE Computational Intelligence Magazine}, 4:77--91, 2009.

\bibitem{KnoCor2003emo}
Joshua~D. Knowles and David Corne.
\newblock Instance generators and test suites for the multiobjective quadratic
  assignment problem.
\newblock In Carlos~M. Fonseca, Peter~J. Fleming, Eckart Zitzler, Kalyanmoy
  Deb, and Lothar Thiele, editors, {\em Evolutionary Multi-criterion
  Optimization, EMO 2003}, volume 2632 of {\em Lecture Notes in Computer
  Science}, pages 295--310. Springer, Heidelberg, Germany, 2003.

\bibitem{lewis2009asynchronous}
Andrew Lewis, Sanaz Mostaghim, and Ian Scriven.
\newblock Asynchronous multi-objective optimisation in unreliable distributed
  environments.
\newblock In {\em Biologically-Inspired Optimisation Methods}, pages 51--78.
  Springer, 2009.

\bibitem{little2019statistical}
Roderick~JA Little and Donald~B Rubin.
\newblock {\em Statistical analysis with missing data}, volume 793.
\newblock John Wiley \& Sons, 2019.

\bibitem{PERT1959}
D.~G. Malcolm, J.~H. Roseboom, C.~E. Clark, and W.~Fazar.
\newblock Application of a technique for research and development program
  evaluation.
\newblock {\em Operations Research}, 7(5):646--669, 1959.

\bibitem{MerBisTraPreuWeiRud11:gecco}
Olaf Mersmann, Bernd Bischl, Heike Trautmann, Mike Preuss, Claus Weihs, and
  G{\"u}nther Rudolph.
\newblock Exploratory landscape analysis.
\newblock In Natalio Krasnogor and Pier~Luca Lanzi, editors, {\em Proceedings
  of the Genetic and Evolutionary Computation Conference, GECCO 2011}, pages
  829--836. ACM Press, New York, NY, 2011.

\bibitem{ohagan2005closed}
Steve O'Hagan, Warwick~B Dunn, Marie Brown, Joshua~D Knowles, and Douglas~B
  Kell.
\newblock Closed-loop, multiobjective optimization of analytical
  instrumentation: gas chromatography/time-of-flight mass spectrometry of the
  metabolomes of human serum and of yeast fermentations.
\newblock {\em Analytical Chemistry}, 77(1):290--303, 2005.

\bibitem{ohagan2007closed}
Steve O'Hagan, Warwick~B Dunn, Joshua~D Knowles, David Broadhurst, Rebecca
  Williams, Jason~J Ashworth, Maureen Cameron, and Douglas~B Kell.
\newblock Closed-loop, multiobjective optimization of two-dimensional gas
  chromatography/mass spectrometry for serum metabolomics.
\newblock {\em Analytical Chemistry}, 79(2):464--476, 2007.

\bibitem{OrseauArmstrong2016}
Laurent Orseau and Stuart Armstrong.
\newblock Safely interruptible agents.
\newblock In {\em Proceedings of the Thirty-Second Conference on Uncertainty in
  Artificial Intelligence}, pages 557--566. AUAI Press, 2016.

\bibitem{pan2009survey}
Sinno~Jialin Pan and Qiang Yang.
\newblock A survey on transfer learning.
\newblock {\em IEEE Transactions on knowledge and data engineering},
  22(10):1345--1359, 2009.

\bibitem{PaqStu06:mqap}
Lu{\'\i}s Paquete and Thomas St{\"u}tzle.
\newblock A study of stochastic local search algorithms for the biobjective
  {QAP} with correlated flow matrices.
\newblock {\em European Journal of Operational Research}, 169(3):943--959,
  2006.

\bibitem{platt2009aptamer}
Mark Platt, William Rowe, David~C Wedge, Douglas~B Kell, Joshua Knowles, and
  Philip~JR Day.
\newblock Aptamer evolution for array-based diagnostics.
\newblock {\em Analytical Biochemistry}, 390(2):203--205, 2009.

\bibitem{PurFle2007tec}
Robin~C. Purshouse and Peter~J. Fleming.
\newblock On the evolutionary optimization of many conflicting objectives.
\newblock {\em IEEE Transactions on Evolutionary Computation}, 11(6):770--784,
  2007.

\bibitem{scott2015evaluation}
Eric~O Scott and Kenneth~A De~Jong.
\newblock Evaluation-time bias in asynchronous evolutionary algorithms.
\newblock In {\em Proceedings of the Companion Publication of the 2015 Annual
  Conference on Genetic and Evolutionary Computation}, pages 1209--1212, 2015.

\bibitem{SnoLarAda2012nips}
Jasper Snoek, Hugo Larochelle, and Ryan~P. Adams.
\newblock Practical {B}ayesian optimization of machine learning algorithms.
\newblock In Peter~L. Bartlett, Fernando C.~N. Pereira, Christopher J.~C.
  Burges, L{\'{e}}on Bottou, and Kilian~Q. Weinberger, editors, {\em Advances
  in Neural Information Processing Systems (NIPS 25)}, pages 2960--2968. Curran
  Associates, Red Hook, NY, 2012.

\bibitem{tabatabaei2015survey}
Mohammad Tabatabaei, Jussi Hakanen, Markus Hartikainen, Kaisa Miettinen, and
  Karthik Sindhya.
\newblock A survey on handling computationally expensive multiobjective
  optimization problems using surrogates: non-nature inspired methods.
\newblock {\em Structural and Multidisciplinary Optimization}, 52(1):1--25,
  2015.

\bibitem{talbi2008parallel}
El-Ghazali Talbi, Sanaz Mostaghim, Tatsuya Okabe, Hisao Ishibuchi, G{\"u}nter
  Rudolph, and Carlos A~Coello Coello.
\newblock Parallel approaches for multiobjective optimization.
\newblock In {\em Multiobjective Optimization}, pages 349--372. Springer, 2008.

\bibitem{thomann2019trust_PhD}
Jana Thomann.
\newblock {\em A trust region approach for multi-objective heterogeneous
  optimization}.
\newblock PhD thesis, Technische Universit\"{a}t Ilmenau, Ilmenau, Germany,
  2019.

\bibitem{THOMANN2019104103}
Jana Thomann and Gabriele Eichfelder.
\newblock Numerical results for the multiobjective trust region algorithm
  {MHT}.
\newblock {\em Data in Brief}, 25:104103, 2019.

\bibitem{thomann2019representation}
Jana Thomann and Gabriele Eichfelder.
\newblock Representation of the {P}areto front for heterogeneous
  multi-objective optimization.
\newblock {\em Journal of Applied and Numerical Optimization}, 1(3):293--323,
  2019.

\bibitem{thomann2019trust}
Jana Thomann and Gabriele Eichfelder.
\newblock A trust-region algorithm for heterogeneous multiobjective
  optimization.
\newblock {\em SIAM Journal on Optimization}, 29(2):1017--1047, 2019.

\bibitem{van2009dimensionality}
Laurens Van Der~Maaten, Eric Postma, and Jaap Van~den Herik.
\newblock Dimensionality reduction: a comparative review.
\newblock {\em J Mach Learn Res}, 10(66-71):13, 2009.

\bibitem{wang2020adaptive}
Xilu Wang, Yaochu Jin, Sebastian Schmitt, and Markus Olhofer.
\newblock An adaptive {B}ayesian approach to surrogate-assisted evolutionary
  multi-objective optimization.
\newblock {\em Information Sciences}, 519:317--331, 2020.

\bibitem{10.1145/3377930.3390147}
Xilu Wang, Yaochu Jin, Sebastian Schmitt, and Markus Olhofer.
\newblock Transfer learning for gaussian process assisted evolutionary
  bi-objective optimization for objectives with different evaluation times.
\newblock In {\em Proceedings of the 2020 Genetic and Evolutionary Computation
  Conference}, page 587–594, New York, NY, USA, 2020. Association for
  Computing Machinery.

\bibitem{yagoubi2011asynchronous}
Mouadh Yagoubi, Ludovic Thobois, and Marc Schoenauer.
\newblock Asynchronous evolutionary multi-objective algorithms with
  heterogeneous evaluation costs.
\newblock In {\em 2011 IEEE Congress of Evolutionary Computation (CEC)}, pages
  21--28. IEEE, 2011.

\end{thebibliography}

\end{document}